\documentclass[10pt,twocolumn]{article}

\usepackage[pagenumbers]{cvpr} 
\usepackage{comment}
\usepackage{tabularx}
\usepackage[most]{tcolorbox}
\usepackage{amsmath}
\usepackage{multirow}
\usepackage{array}
\usepackage{caption}
\usepackage{wrapfig}
\usepackage{enumitem}
\usepackage{tikz}
\usepackage{lipsum}
\usepackage{subcaption}
\usepackage{multirow} 
\usepackage{adjustbox}
\usepackage[table]{xcolor}  
\usepackage{siunitx}          
\usepackage{graphicx}     
\usepackage{amssymb}
\usepackage[utf8]{inputenc}
\usepackage{float} 

\usepackage{booktabs}
\usepackage[table]{xcolor}
\usepackage{siunitx}
\usepackage[margin=1in]{geometry}
\usepackage{microtype}

\newcommand{\pos}[1]{\cellcolor{green!15}{#1}}
\newcommand{\nat}[1]{\cellcolor{white}{#1}}
\renewcommand{\neg}[1]{\cellcolor{red!12}{#1}}
\newcommand{\cyan}[1]{\cellcolor{cyan!20}#1}
\newcommand{\cyantext}[1]{\textcolor{cyan!50}{#1}}

\newcommand{\postext}[1]{\textcolor{green!40}{#1}}
\newcommand{\negtext}[1]{\textcolor{red!32}{#1}}

\renewcommand{\paragraph}[1]{\vspace{1.25mm}\noindent\textbf{#1}}
\def\jh#1{{\bf [Junlin:} {\it\color{green} {#1}}{\bf ]}.}

\definecolor{cvprblue}{rgb}{0.21,0.49,0.74}
\usepackage[pagebackref,breaklinks,colorlinks,allcolors=cvprblue]{hyperref}

\title{
From Pixels to Feelings: Aligning MLLMs with Human Cognitive Perception of Images}

\author{
{\normalsize
Yiming Chen$^{1*}$,\ 
Junlin Han$^{1*}$,\ 
Tianyi Bai$^{2}$,\ 
Shengbang Tong$^{4}$,\ 
Filippos Kokkinos$^{3}$,
Philip Torr$^{1}$\ 
}\\[0.7em]
{\small
$^{1}$Oxford University,\ 
$^{2}$HKUST,\ 
$^{3}$University College London,\ 
$^{4}$New York University
}\\[0.9em]
{\small
* Equal contribution.\quad
}\\[0.4em]
{\small
Project page: https://follen-cry.github.io/MLLM-Cognition-project-page/
}
}

\begin{document}
\maketitle
\begin{figure*}[t]
    \centering
    \includegraphics[width=\linewidth]{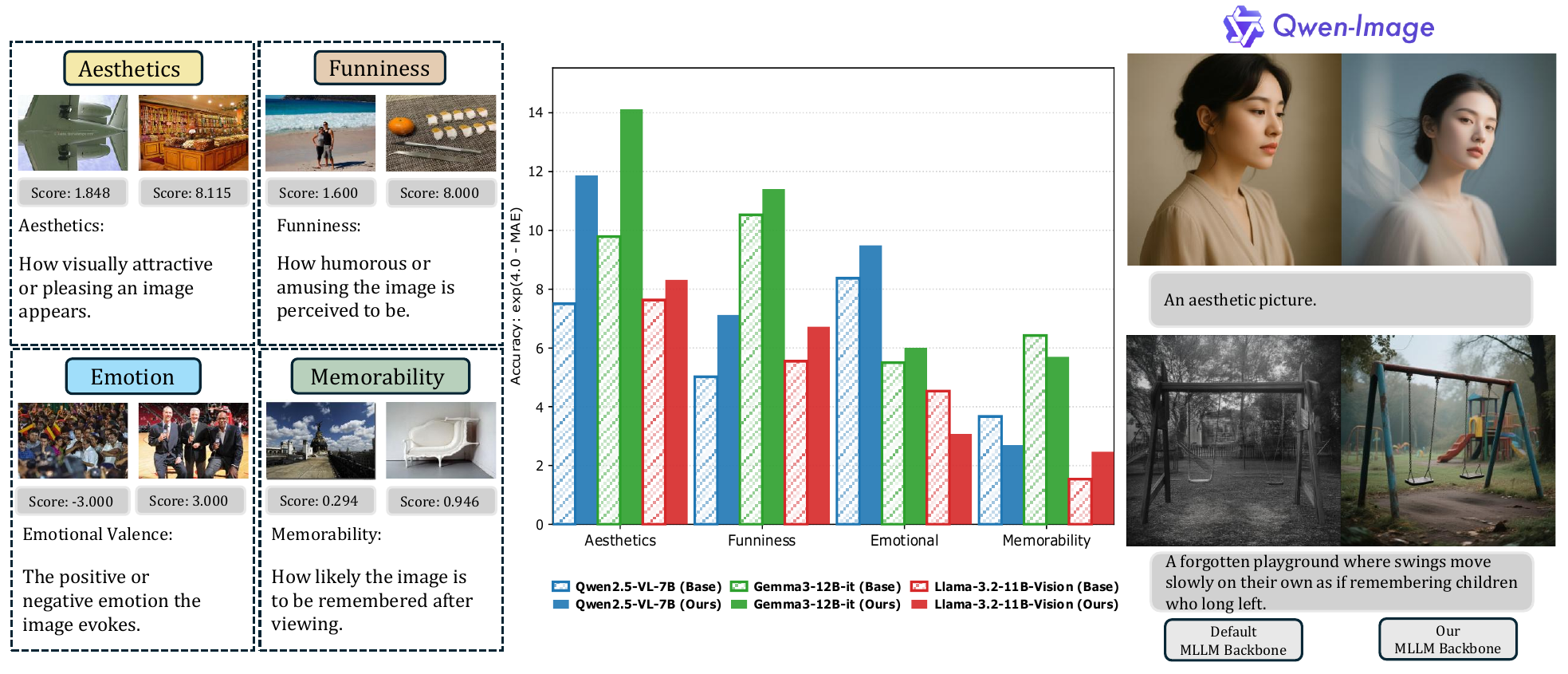}
    \caption{We present CogIP-Bench, a comprehensive cognition benchmark that evaluates the alignment of cognition score prediction between MLLM and humans. \textbf{Left:} example datapoints for each dimension: aesthetics, funniness, emotion and memorability. \textbf{Middle:} post-training results of three popular MLLMs across different dimensions. \textbf{Right:} results of swapping the MLLM backbone, comparing the effect of cognition-related image generation with the Qwen-Image pipeline.}
    \label{fig:cognition_benchmark}
\end{figure*}

\begin{abstract}
While Multimodal Large Language Models (MLLMs) are adept at answering ``what" is in an image—identifying objects and describing scenes—they often lack the ability to understand ``how" an image feels to a human observer. This gap is most evident when considering subjective cognitive properties, such as what makes an image memorable, funny, aesthetically pleasing, or emotionally evocative. To systematically address this challenge, we introduce CogIP-Bench, a comprehensive benchmark for evaluating MLLMs on such image cognitive properties. Our evaluation reveals a significant gap: current models are poorly aligned with human perception of these nuanced properties. We then demonstrate that a post-training phase can effectively bridge this gap, significantly enhancing the model's alignment with human judgments. Furthermore, we show that this learned cognitive alignment is not merely predictive but also transferable to downstream creative tasks. By integrating our cognitively-aligned MLLM into an image generation pipeline, we can guide the synthesis process to produce images that better embody desired traits, such as being more memorable or visually appealing. Our work provides a benchmark to measure this human-like perception, a post-training pipeline to enhance it, and a demonstration that this alignment unlocks more human-centric AI.

\end{abstract}

\section{Introduction}

Advances in MLLMs, such as GPT \cite{openai_gpt-4o_2024} and Gemini \cite{team_gemini_2025}, have led to significant breakthroughs in visual understanding. Their success in tasks like visual question answering and detailed image captioning demonstrates a strong ability to perform objective recognition and description \cite{Zhou2019UnifiedVL, AIP2024VQASurvey,alayrac2022flamingovisuallanguagemodel}. Yet, despite this progress, a crucial dimension of visual intelligence remains largely unexplored: the perception of subjective, human-centered cognitive properties~\cite{goetschalckx2019ganalyzevisualdefinitionscognitive, isola_what_2014, Valenzise2022}. These properties—encompassing aesthetics, humor, emotion, and memorability—are fundamental to the human visual experience but pose a challenge for current models, which are primarily trained on factual, descriptive data.

Previous research has explored individual cognitive image properties, developing models to predict the memorability of an image \cite{han_what_2023,ICCV15_Khosla,isola_what_2014}, quantify its aesthetic appeal \cite{laion_aesthetic_predictor, liu_unlocking_2025,ren2017personalized}, or even assess its potential for humor \cite{li_oxfordtvg-hic_2023,jain2025humordbaiunderstandgraphical,hessel2023androids}. 
Although valuable, these efforts often focus on specialized, vision-only models or statistical analysis, leaving a significant gap: there is no systematic framework to evaluate and enhance these comprehensive cognitive abilities within powerful, general-purpose MLLMs. It remains unclear how well MLLMs align with human judgments on aesthetics, funniness, emotional valence, and memorability; whether they can be taught to acquire this sophisticated understanding; and what the downstream effects would be once they are more aligned with humans.

To bridge this gap, we introduce the CogIP-Bench, the first comprehensive benchmark designed to systematically measure the cognitive alignment between MLLMs and humans in four key dimensions. Our extensive evaluation of leading MLLMs in CogIP-Bench reveals a significant discrepancy, confirming that these models, despite their advanced capabilities, are poorly aligned with human perception of these subjective traits. For example, all MLLMs, including open-source and API-based models, showed nearly 0 correlation with humans on the memorability level of images, while the average Spearman correlation in other cognitive dimensions remains below 0.5. (Table \ref{tab:CogIP-BenchResults})

Beyond merely identifying this problem and benchmarking, we demonstrate a practical solution. We show that a targeted post-training pipeline with training data in CogIP-Bench can effectively instill this cognitive knowledge into an MLLM, significantly improving its ability to predict human cognitive scores. One more compelling finding is that this newly acquired cognitive alignment is not just a passive predictive skill, but a transferable capability. By integrating our fine-tuned MLLM as the MLLM backbone in a state-of-the-art image generation pipeline (Qwen-Image)  \cite{wu2025qwenimagetechnicalreport}, we show that it can guide the synthesis process to create images that are intentionally more memorable, aesthetically pleasing, or emotionally resonant.

Our main contributions can be summarized as follows:

\begin{itemize}

\item \textbf{Benchmark:} We introduce CogIP-Bench, a novel benchmark to evaluate MLLMs on four fundamental cognitive image properties: aesthetics, funniness, emotional valence, and memorability.

\item \textbf{Alignment:} We present a post-training pipeline and data that significantly improves an MLLM's alignment with human cognitive judgments, effectively teaching it a more human-like ``sense" of these properties.

\item \textbf{Transferability:} We provide the first demonstration that enhancing cognitive alignment in an MLLM can directly translate into more controllable and human-centric outcomes in downstream generative tasks, such as text-to-image synthesis.

\end{itemize}

\section{Related Work}

\begin{figure*}[t]
    \centering
    \includegraphics[width=\linewidth]{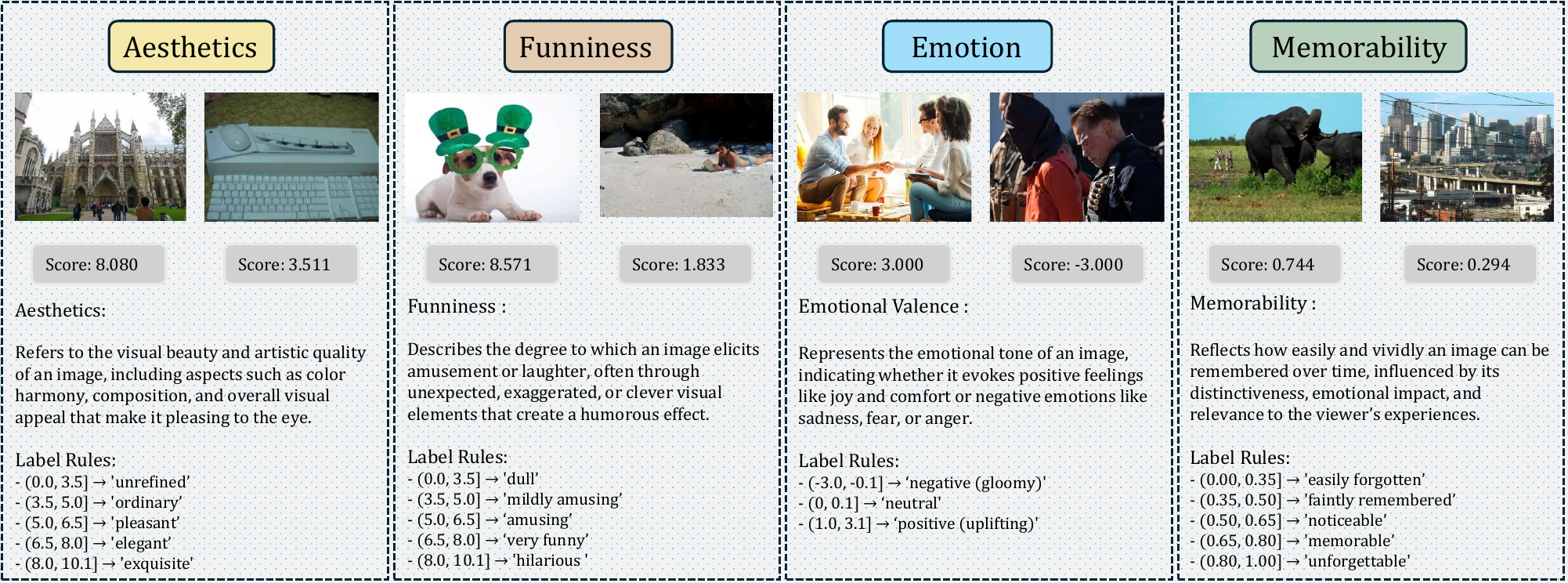}
    \caption{Examples of the CogIP-Bench, for each cognition dimension, we show two images along with their cognition scores and the interpretation of that cognitive dimension. }
    \label{fig:cognition_benchmark}
\end{figure*}

\subsection{MLLMs and Their Evaluations}

\textbf{Multimodal large language models (MLLM)}, \cite{wang2024qwen2vlenhancingvisionlanguagemodels,qwen2025qwen25technicalreport,yang2025qwen3technicalreport,team_gemini_2025,li_llava-onevision_2024,openai_gpt-4o_2024,liu_visual_2023,liu_improved_2024,tong_cambrian-1_2024, Yin_2024, caffagni2024revolutionmultimodallargelanguage, zhang2024mmllmsrecentadvancesmultimodal, zhang2025crossmodalinformationflowmultimodal, chen2024multimodallargelanguagemodels} combine visual encoders with pretrained language models, allowing them to reason jointly over text and images. These systems often use visual encoders such as CLIP \cite{radford2021learningtransferablevisualmodels}, along with adapter mechanisms—such as MLPs, query-based transformers, or attention layers—to bridge the two modalities. They demonstrate promising application values in a broad fields \cite{bai_digirl_2024,chu_sft_2025,tong_metamorph_2024,wang_mllm-tool_2025,xu_cad-mllm_2025,zhai_fine-tuning_2024,zhou_transfusion_2024,driess2023palmeembodiedmultimodallanguage} such as CAD drawing, robotics agentic system.
\\
\textbf{MLLM Benchmarks}. With the rapid advancement of MLLM, all-round benchmarks have been proposed to assess diverse capabilities of MLLMs \cite{liu2024mmbench,yue2023mmmu,fu2023mme,li2023seed,fu2024ocrbenchv2improvedbenchmark,lu2024mathvista,mathew2021docvqa,kembhavi2016diagram,bai2025hallucination, li2024surveybenchmarksmultimodallarge}. They tested a wide range of capabilities that require both visual and textual modalities as input, ranging from text-visual grounding, perception, optical character recognition and compositional reasoning. However, there lacks a benchmark that assess the cognitive ability and the degree of alignment, in subjective properties like aesthetics, funniness, emotional valence and memorability, between MLLMs and human. Hence, our CogIP-Bench provides a pipeline for assessing how similar MLLMs perceive an image with humans from a cognitive perspective.

\subsection{Human Cognition in AI}

Research on the cognitive abilities of AI models has explored both the alignment of decision-making between humans and models \cite{xie_can_2024, binz_foundation_2025, gui_challenge_2023} and the interpretation of mechanisms of neural network models from neuroscientific and psychological perspectives \cite{song_decoding_2023, ji-an_discovering_2025, singh_eeg2image_2023}. 
Before LLMs, some studies also demonstrated the abilities of AI models in cognitive properties of images; for example, research \cite{han_what_2023, isola_what_2014} focus on the memorability of images trying many different AI models, from linear classifiers to ViTs in order to quantify and understand it. GANalyze \cite{goetschalckx2019ganalyzevisualdefinitionscognitive} use Generative Adversarial Networks (GANs) to study the aesthetics, memorability, and emotional valence of images. 

However, most existing studies primarily focus on text-only large language models \cite{binz_foundation_2025, xie_can_2024}, vision-only models \cite{song_decoding_2023}, or simpler recurrent neural networks \cite{ji-an_discovering_2025}. Recent work has begun to assess the visual cognition abilities of MLLMs \cite{buschoff_visual_2024}, though such analysis are largely behavioral—centering on spatial reasoning and causal inference tasks. In contrast, the cognition abilities examined by our CogIP-Bench emphasize the subjective experience elicited by visual stimuli, assessing the alignment between MLLMs and human in terms of aesthetics, funniness, emotion and memorability properties of images. While prior studies have examined individual cognitive properties, some of which adopt statistical approaches \cite{jang2022modelingquantifyingpredictingsubjectivity}, others concentrating on vision-only models \cite{li_oxfordtvg-hic_2023, jain2025humordbaiunderstandgraphical, jing_unforgettable_2025, han_what_2023}, and still others evaluating the cognitive abilities of MLLMs \cite{mertens2024findingemoimagedatasetemotion, liu_unlocking_2025}, these efforts typically address only a single dimension of cognition rather than providing a comprehensive perspective.
Our work is the first systematic attempt to construct a benchmark that focuses on comprehensive subjective cognition (covering 4 dimensions) on images, offering new insights into MLLM cognitive evaluation.

\section{CogIP-Bench}

\begin{table*}[h]
\centering
\small
\setlength{\tabcolsep}{4pt}
\renewcommand{\arraystretch}{1.15}

\begin{adjustbox}{width=\textwidth}
\begin{tabular}{
l
S[table-format=1.3] S[table-format=1.3] S[table-format=2.3]
S[table-format=1.3] S[table-format=1.3] S[table-format=2.3]
S[table-format=1.3] S[table-format=1.3] S[table-format=2.3]
S[table-format=1.3] S[table-format=1.3] S[table-format=2.3]
}
\toprule
\textbf{Model} &
\multicolumn{3}{c}{\textbf{Aesthetics}} &
\multicolumn{3}{c}{\textbf{Funniness}} &
\multicolumn{3}{c}{\textbf{Emotional}} &
\multicolumn{3}{c}{\textbf{Memorability}} \\
\cmidrule(lr){2-4} \cmidrule(lr){5-7} \cmidrule(lr){8-10} \cmidrule(lr){11-13}
 & \textbf{MSE} & \textbf{MAE} & \textbf{Spearman} & \textbf{MSE} & \textbf{MAE} & \textbf{Spearman} & \textbf{MSE} & \textbf{MAE} & \textbf{Spearman} & \textbf{MSE} & \textbf{MAE} & \textbf{Spearman} \\
\midrule
Specific Model        
& 0 & 0 & 1 
& 3.420 & 1.470 & 0.581 
& 4.919 & 1.782 & 0.585 
& 7.800 & 1.650 & 0.504 \\

\midrule
\multicolumn{13}{c}{\textbf{Open-source Models}} \\
\midrule

Qwen2-VL-7B-Instruct
& 4.491 & 1.672 & 0.507
& 9.534 & 2.442 & 0.440
& 9.289 & 2.240 & 0.505
& 9.100 & 2.250 & \neg{-0.019} \\

Qwen2.5-VL-7B-Instruct
& 5.680 & 1.981 & 0.533
& 8.073 & 2.151 & 0.485
& 9.433 & 2.255 & 0.537
& 10.200 & 2.350 & 0.096 \\

Qwen3-VL-8B-Instruct
& \neg{6.892} & \neg{2.141} & 0.457
& 9.012 & 2.284 & 0.513
& 11.756 & 2.681 & 0.545
& 10.200 & 2.430 & 0.033 \\

Gemma3-12B-it (Base)
& 4.349 & 1.719 & 0.478
& \pos{4.716} & \pos{1.646} & \pos{0.651}
& 9.156 & 2.295 & \pos{0.588}
& 8.000 & \pos{2.140} & 0.040 \\

llava-v1.6-mistral-7b-hf
& \pos{2.202} & \pos{1.119} & \pos{0.595}
& 9.872 & 2.400 & 0.257
& 7.071 & 2.127 & 0.380
& \neg{28.900} & \neg{4.700} & 0.056 \\

llava-v1.6-vicuna-7b-hf
& 4.112 & 1.681 & \neg{0.346}
& \neg{10.772} & 2.524 & \neg{-0.147}
& 9.375 & 2.237 & 0.468
& \pos{6.500} & 2.180 & 0.013 \\

llava-v1.6-vicuna-13b-hf
& 2.896 & 1.347 & 0.510
& 9.389 & 2.331 & 0.053
& \pos{6.817} & \pos{2.024} & 0.467
& 11.600 & 2.680 & 0.017 \\

llava-onevision-qwen2-7b-ov-hf
& 5.861 & 1.983 & 0.436
& 10.311 & \neg{2.555} & 0.344
& 11.619 & 2.683 & 0.558
& 10.200 & 2.340 & -0.013 \\

Llama-3.2-11B-Vision-Instruct
& 4.885 & 1.823 & 0.427
& 6.284 & 1.924 & 0.404
& \neg{12.159} & \neg{2.796} & \neg{0.335}
& 9.700 & 2.550 & \pos{0.135} \\

\midrule
\multicolumn{13}{c}{\textbf{API-based Models}} \\
\midrule
GPT-4o
& \pos{4.073} & \pos{1.597} & 0.485
& 8.112 & 2.151 & 0.565
& 6.667 & 1.922 & 0.658
& \pos{7.630} & 2.234 & 0.057 \\

GPT-5
& 4.896 & 1.792 & 0.459
& \pos{3.262} & \pos{1.342} & \pos{0.731}
& \pos{6.205} & \pos{1.902} & 0.657
& 8.500 & \pos{2.128} & \pos{0.115} \\

Claude-Haiku-4.5
& 4.884 & 1.806 & \pos{0.490}
& \neg{9.982} & \neg{2.403} & \neg{0.431}
& \neg{11.999} & \neg{2.704} & \neg{0.503}
& 8.120 & 2.177 & \pos{0.115} \\

Gemini-2.5-Pro
& \neg{6.061} & \neg{1.984} & \neg{0.201}
& 5.125 & 1.819 & 0.686
& 6.471 & 1.926 & \pos{0.726}
& \neg{11.300} & \neg{2.588} & \neg{-0.036} \\

\bottomrule
\end{tabular}
\end{adjustbox}

\caption{Test results of various open and closed source MLLMs on our \textbf{CogIP-Bench}. Each column pair shows MSE and MAE for each cognitive subtask (Aesthetics, Funniness, Emotional, Memorability). The dimension-specific models are vision-only models trained to predict the cognition scores of the image for each dimension, serve as a reference. \postext{Green} entry color indicates the \postext{best} performance under each metric column (separately for open-scource and API-based models), and \negtext{red} indicates the \negtext{worst}.}
\label{tab:CogIP-BenchResults}
\end{table*}

Our benchmark is designed to test the cognition alignment of MLLM with human, in terms of score prediction on image. So each data-point contains(1) an image for perceiving, (2) a prompt to provide specific cognition context and instruct the model to predict cognition score and (3) a human preference score as ground truth. 

\subsection{Benchmark Construction}
Our benchmark contains 4 dimensions: Aesthetics, Funniness, Emotional Valence and Memorability. The collection pipeline is listed as follows: 


\textbf{Aesthetics}: In the context of images, aesthetics capture the perceptual and affective qualities that evoke visual appeal, harmony, and artistic value in human observers. For this dimension, we draw image data from the LaMem dataset \cite{ICCV15_Khosla}, while ground-truth aesthetic scores are provided by the LAION-Aesthetic-Predictor-V1 model \cite{laion_aesthetic_predictor}. This predictor employs a CLIP vision encoder with a lightweight regression head, trained to estimate aesthetic quality and output continuous scores. Its training is based on annotated ratings collected from a photo-community website, ensuring alignment with human judgments of visual appeal~\cite{Murray2012_AVA}.

\textbf{Funniness}: Humor is widely accepted by ``incongruity theory", which suggests that it arises from unexpected subversions of contextual expectations \cite{Attardo2020}. However, we are doing an unimodal analysis (only assess image without caption) which is inherently a ``stateless" process, that is, no presummed expectation of the context, hence we call this cognition dimension funniness. We use the HumorDB dataset \cite{jain2025humordbaiunderstandgraphical}, which is a carefully curated dataset designed to evaluate and advance visual humor understanding by AI systems. It comprises a diverse range of images from photos, cartoons, sketches, and AI-generated content all annotated with a humorous score, which is collected by crowdsourcing in three annotation tasks: binary classification, range rating and pairwise comparison.

\textbf{Emotional Valence}: We use the FindingEmo dataset \cite{mertens2024findingemoimagedatasetemotion}, which is a dataset focusing on emotional understanding of MLLM and containing diverse image-text pairs annotated with human judgements of emotional valence and intensity via online crowdsourcing, where emotional valence is a bi-directional metric that emphasizes on the positive or negative quality of an emotion rather than the arousal level of it.  It aims to test a model's ability to perceive, interpret, and align visual cues with emotional semantics.

\textbf{Memorability}: People tend to memorize images with central objects, faces or salient actions better than natural landscapes which suggests that there are certain visual intrinsics or pattern that make some images more memorable than others \cite{han_what_2023, ICCV15_Khosla} despite personal experiences. To investigate this property and alignment with human perception in MLLM, we used the LaMem dataset along with their memorability scores provided \cite{ICCV15_Khosla} as the source dataset for the memorability dimension to construct our benchmark. The LaMem dataset is a large-scale memorability dataset that contains 60,000 images from diverse sources and the scores are derived from reliable and large-scale human experiments with a carefully designed visual memory game.

For each cognition dimension except emotional valence, when sampling images, we define 5 even score ranges, i.e., (0, 2), (2, 4), (4, 6), (6, 8), (8, 10) where the raw scales of aesthetics and funniness are 0-10 but memorability is 0-1 (we scale it to 1-10 accordingly). However, due to the bi-directional nature of emotional valence, its raw score ranges from -3 to 3 hence we split it into 3 buckets: (-3, -1), (-1, 1) and (1, 2) and later scale it to 1-10 when computing the MSE and MAE of test results. We then enforce the same number of datapoints sampled in each bucket to ensure the balance of data in terms of human preferences score.

\subsection{Benchmark Details}
For each cognition property, we collect 800 datapoints for training and 120 for test. Each datapoint is composed of an image, a cognition-specific prompt and the ground truth answer which includes the numeric score. 
The prompt is carefully curated to help the MLLM understand the meaning of each particular cognitive property. More details are provided in the \textit{supplementary material}. Some examples are shown in Figure \ref{fig:cognition_benchmark}.


\section{Benchmarking Results}
We evaluate the effectiveness of our CogIP-Bench in aligning subjective cognition on images between humans and MLLMs.
 In Section \ref{subsec:experiment settings}, we outlined our experimental settings. Section \ref{subsect:result on CogIP-Bench} presents the results from multiple models on our CogIP-Bench and  Section \ref{subsec:method} introduces our post-training pipeline for injecting cognitive image knowledge into MLLMs. In Section \ref{subsec:other benchmark after sft}, we show the effect of post-training and the impact on other benchmarks.

\subsection{Experimental Settings}\label{subsec:experiment settings}

We test open-source as well as API-based MLLMs on our benchmark to evaluate their alignment with human cognition.  These open-source models include: Qwen2-VL-7B \cite{wang2024qwen2vlenhancingvisionlanguagemodels}, Qwen2.5-VL-7B \cite{qwen2025qwen25technicalreport}, Qwen3-VL-8B \cite{yang2025qwen3technicalreport}, Gemma3-12B-it \cite{team_gemini_2025}, llava-v1.6-mistral-7b-hf, llava-v1.6-vicuna-7b-hf, llava-v1.6-vicuna-13b-hf \cite{liu_improved_2024}, llava-onevision-qwen2-7b-ov-hf \cite{li_llava-onevision_2024}, and Llama-3.2-11B-Vision-Instruct \cite{grattafiori2024llama3herdmodels}. 
The API-based models are: GPT-4o \cite{openai_gpt-4o_2024}, GPT-5 \cite{openai_gpt5_systemcard_2025}, Claude-Haiku-4.5 \cite{anthropic_claude_sonnet45_systemcard_2025} and Gemini-2.5-Pro \cite{team_gemini_2025}. 

We also report the results of specific models for each cognition dimension to provide comparisons.\\
\textbf{Aesthetics}: We used LAION aesthetic predictor \cite{laion_aesthetic_predictor} to get the score, which is the ground truth predictor by definition.\\
\textbf{Funniness}: We trained a regressor model with a CLIP vision encoder and a linear projector as the output layer. We used the images and ground-truth score of our own CogIP-Bench for training. \\
\textbf{Emotional Valence}: same as Funniness.\\
\textbf{Memorability}: We used the off-the-shelf HumanMem predictor model \cite{han_what_2023}.

\begin{table*}[t]
\centering
\small
\setlength{\tabcolsep}{4pt}
\renewcommand{\arraystretch}{1.15}

\begin{adjustbox}{width=\textwidth}
\begin{tabular}{
l
S[table-format=1.3] S[table-format=1.3] S[table-format=2.3]
S[table-format=1.3] S[table-format=1.3] S[table-format=2.3]
S[table-format=1.3] S[table-format=1.3] S[table-format=2.3]
S[table-format=1.3] S[table-format=1.3] S[table-format=2.3]
}
\toprule
\textbf{Model} &
\multicolumn{3}{c}{\textbf{Aesthetics}} &
\multicolumn{3}{c}{\textbf{Funniness}} &
\multicolumn{3}{c}{\textbf{Emotional}} &
\multicolumn{3}{c}{\textbf{Memorability}} \\
\cmidrule(lr){2-4} \cmidrule(lr){5-7} \cmidrule(lr){8-10} \cmidrule(lr){11-13}
 & \textbf{MSE} & \textbf{MAE} & \textbf{Spearman} & \textbf{MSE} & \textbf{MAE} & \textbf{Spearman} & \textbf{MSE} & \textbf{MAE} & \textbf{Spearman} & \textbf{MSE} & \textbf{MAE} & \textbf{Spearman} \\
\midrule
Specific Model
& 0 & 0 & 1
& 3.420 & 1.470 & 0.581
& 4.919 & 1.782 & 0.585
& 7.800 & 1.650 & 0.504 \\
\midrule
Qwen2.5-VL-7B (base)
& 5.669 & 1.985 & 0.400
& 9.376 & 2.387 & 0.486
& 6.378 & 1.875 & 0.635
& 13.200 & 2.700 & 0.086 \\

Qwen2.5-VL-7B (SFT)
& \cyan{3.841} & \cyan{1.526} & \cyan{0.450}
& \cyan{7.022} & \cyan{2.037} & \cyan{0.509}
& \cyan{5.775} & \cyan{1.750} & \cyan{0.644}
& 15.500 & 3.010 & 0.012 \\

Gemma3-12B-it (Base)
& 4.349 & 1.719 & 0.478
& 4.716 & 1.646 & 0.651
& 9.156 & 2.295 & 0.588
& 8.000 & 2.140 & 0.040 \\

Gemma3-12B-it (SFT)
& \cyan{2.614} & \cyan{1.353} & 0.473
& \cyan{3.986} & \cyan{1.566} & \cyan{0.654}
& \cyan{8.208} & \cyan{2.208} & 0.573
& \cyan{7.500} & 2.260 & \cyan{0.165} \\

Llama-3.2-11B-Vision-Instruct (Base)
& 5.980 & 1.968 & 0.449
& 8.922 & 2.286 & 0.298
& 9.760 & 2.489 & 0.531
& 47.680 & 3.570 & -0.036 \\

Llama-3.2-11B-Vision-Instruct (SFT)
& \cyan{5.783} & \cyan{1.882} & 0.387
& \cyan{7.121} & \cyan{2.095} & \cyan{0.327}
& 12.167 & 2.878 & 0.495
& \cyan{15.410} & \cyan{3.096} & -0.146 \\

\bottomrule
\end{tabular}
\end{adjustbox}

\caption{Test results on \textbf{CogIP-Bench}. Each column pair shows MSE and MAE (after normalization) for a cognitive subtask (Aesthetics, Funniness, Emotional, Memorability). The \cyantext{blue} (\cyantext{better}) color indicates the performance improvement of alignment after SFT of each MLLM for each cognitive dimension metric.}
\label{tab:sft CogIP}
\end{table*}
\subsection{Benchmark Results on MLLMs} \label{subsect:result on CogIP-Bench}
We tested the cognition alignment capability of MLLMs using our own CogIP-Bench (Table \ref{tab:CogIP-BenchResults}). 
We observed that all models fail to capture the memorability of images with near-zero Spearman correlation with human preferences, which indicates that memorability is the most challenging and abstract cognitive property of images among the four cognitiive dimensions. 
For open-source models, the alignment with human cognition appears stronger in aesthetics and emotional valence, whereas the funniness dimension exhibits greater variability in alignment scores. In contrast, API-based models demonstrate relatively weaker alignment on aesthetics, but achieve significantly better alignment on funniness and emotional valence, outperforming open-source counterparts in these dimensions.

Moreover, we find that MLLMs with better performance on other benchmark, such as the Qwen series and Gemma3, also perform relatively well on our CogIP-Bench. For the API-based models, GPT-5 yields better performance than others. This suggests that there is a positive correlation between these subjective human cognitive properties and the general vision-language capabilities.


\subsection{Post-training Method}\label{subsec:method}
We employ standard supervised fine-tuning (SFT) to inject cognitive knowledge into the model. A key challenge in this numerical score prediction task lies in the inherent insensitivity of language models to numerical values. Since numbers are tokenized into discrete tokens in the same manner as words, the model lacks an explicit understanding of the ordinal or distance relationships between them. To address this limitation, we explore two complementary approaches:

First, the prompts are designed to instruct the model to predict a categorical label as a classification task. For each cognition dimension except emotional valence, the labels are: [very low, low, medium, high, very high], while for emotional valence the labels are: [negative, neutral, positive]. Based on the provided label-to-score mapping rule, the model predicts a corresponding numeric score. This two-step formulation enables the model to first provide a qualitative judgment, leveraging its strength in natural language reasoning, and subsequently produce a quantitative score grounded in that classification.

Next, we experimented with the soft-label loss~\cite{Wang2025Enhancing}. 
As mentioned above, the standard cross-entropy loss ($L_{\mathrm{CE}}$) uses one-hot encoding, 
which means that $\mathbf{q}$ is a Dirac distribution 
($q(i)=1$ only when $i=t$), where $t$ is the index of the target token~\eqref{eq:ce_loss}, $V$ is the vocabulary size and $p$ is the probability distribution of the vocabulary ($\mathbf{p} \in \mathbb{R}^V \text{ such that } \sum_{i=1}^{V} p_i = 1.$
).
Hence, the numerical distance information between digit tokens is lost when computing the loss. For example, the numbers 1 and 9 would incur the same loss when the target token is 2, even though 1 is much closer to 2. 

\begin{equation}
L_{\mathrm{CE}}(\mathbf{p}, \mathbf{q}(t)) 
= \sum_{i=1}^{V} q_i(t)\,(-\log p_i),
\label{eq:ce_loss}
\end{equation}

To address this, we switch to a soft-label distribution ($\mathbf{q}^{SL}$) for numerical tokens. 
Instead of a Dirac function, we construct $\mathbf{q}$ (only for numeric target tokens) as shown in ~\eqref{eq:softlabel}, where $\delta$ is the old Dirac distribution and $\eta$ is a mixture coefficient balancing the degree of ``softness" of the new loss function. 
This encourages the model to output digit token that are ``closer” to the target value in a continuous manner. 
We use $\psi(t)$ as a triangular function, as suggested by the original paper~\cite{Wang2025Enhancing}.

\begin{equation}
\mathbf{q}^{SL}(t) = (1 - \eta)\,\delta(t) + \eta\,\psi(t)
\label{eq:softlabel}
\end{equation}

We use LoRA training instead of full fine-tuning to prevent overfitting.  More explorations on using reinforcement learning (RL) for post-training are provided in the discussions (Section \ref{sec:rl}). 

\subsection{Post-training Results} \label{subsec:other benchmark after sft}
In this section, we analyze the effect of SFT as reflected in our CogIP-Bench results, comparing the model performance before and after SFT for three models: Qwen2.5-VL-7B, Gemma3-12B-it, and Llama-3.2-11B, as shown in Table \ref{tab:sft CogIP}. We then examine how this alignment influences other aspects of each model’s capability by evaluating them across a diverse set of benchmarks. Following the common task categorization used in Cambrian-1, Web-SSL, and LSBS~\cite{tong_cambrian-1_2024,fan2025scaling,han2025lsbs}, we group the tasks into four categories: Vision-Centric, OCR, General, and Knowledge (Table \ref{tab:other benchmark}), each comprising three benchmarks \cite{tong2024eyeswideshutexploring,realworldqa2024,zhu2025cvbenchevaluatingcrossvideosynergies,Liu_2024,mathew2021docvqa,masry-etal-2022-chartqa,fu2023mme,liu2024mmbench,li2023seed,yue2023mmmu,lu2024mathvista,kembhavi2016diagram}. More details are provided in the \textit{supplementary material}.

From Table \ref{tab:sft CogIP}, we observe that aside from memorability—where all models exhibit difficulty in prediction—there is a consistent alignment improvement across nearly all cognitive dimensions and metrics. This suggests that SFT effectively incorporates human subjective preferences in image perception into MLLMs, thereby enhancing their alignment with human judgments.

Regarding performance on other benchmarks (Table \ref{tab:other benchmark}), we find that Qwen2.5-VL-7B and Llama-3.2-11B maintain comparable performance with acceptable variance, while Gemma3-12B-it demonstrates substantial improvements across all categories. These results further indicate a positive correlation between subjective cognitive alignment and overall model capability.

\begin{table}[H]
\centering
\small
\setlength{\tabcolsep}{5pt}
\renewcommand{\arraystretch}{1.15}

\begin{adjustbox}{width=0.5\textwidth,center}
\begin{tabular}{lcccccccccccccccc}
\toprule
\textbf{Model} &
\multicolumn{1}{c}{\textbf{Vision-Centric}} &
\multicolumn{1}{c}{\textbf{OCR}} &
\multicolumn{1}{c}{\textbf{General}} &
\multicolumn{1}{c}{\textbf{Knowledge}} \\
\midrule

Qwen2.5-VL-7B-Base &
 0.740 &
 0.898 &
 0.818 &
 0.702 \\

Qwen2.5-VL-7B-SFT &
0.738&
0.896&
\cyan{0.820}&
 \cyan{0.703} \\
\midrule
Gemma3-12B-it-Base &
 0.674 &
 0.599 &
 0.735 &
 0.633 \\

Gemma3-12B-it-SFT &
 \cyan{0.684} &
 \cyan{0.651} &
 \cyan{0.745} &
 \cyan{0.656} \\
\midrule
Llama-3.2-11B-VI-Base &
 0.686 &
 0.790 &
 0.703 &
 0.577 \\

Llama-3.2-11B-VI-SFT &
0.685 &
 \cyan{0.793} &
 0.700 &
 \cyan{0.585} \\
\bottomrule
\end{tabular}
\end{adjustbox}

\caption{Benchmark comparison across multiple models with per-section averages (Vision-Centric, OCR, General, Knowledge).}
\label{tab:other benchmark}
\end{table}

\section{Image Generation Results} \label{sec:qwenimage}
\begin{figure*}[t]
    \centering
    \includegraphics[width=\linewidth]{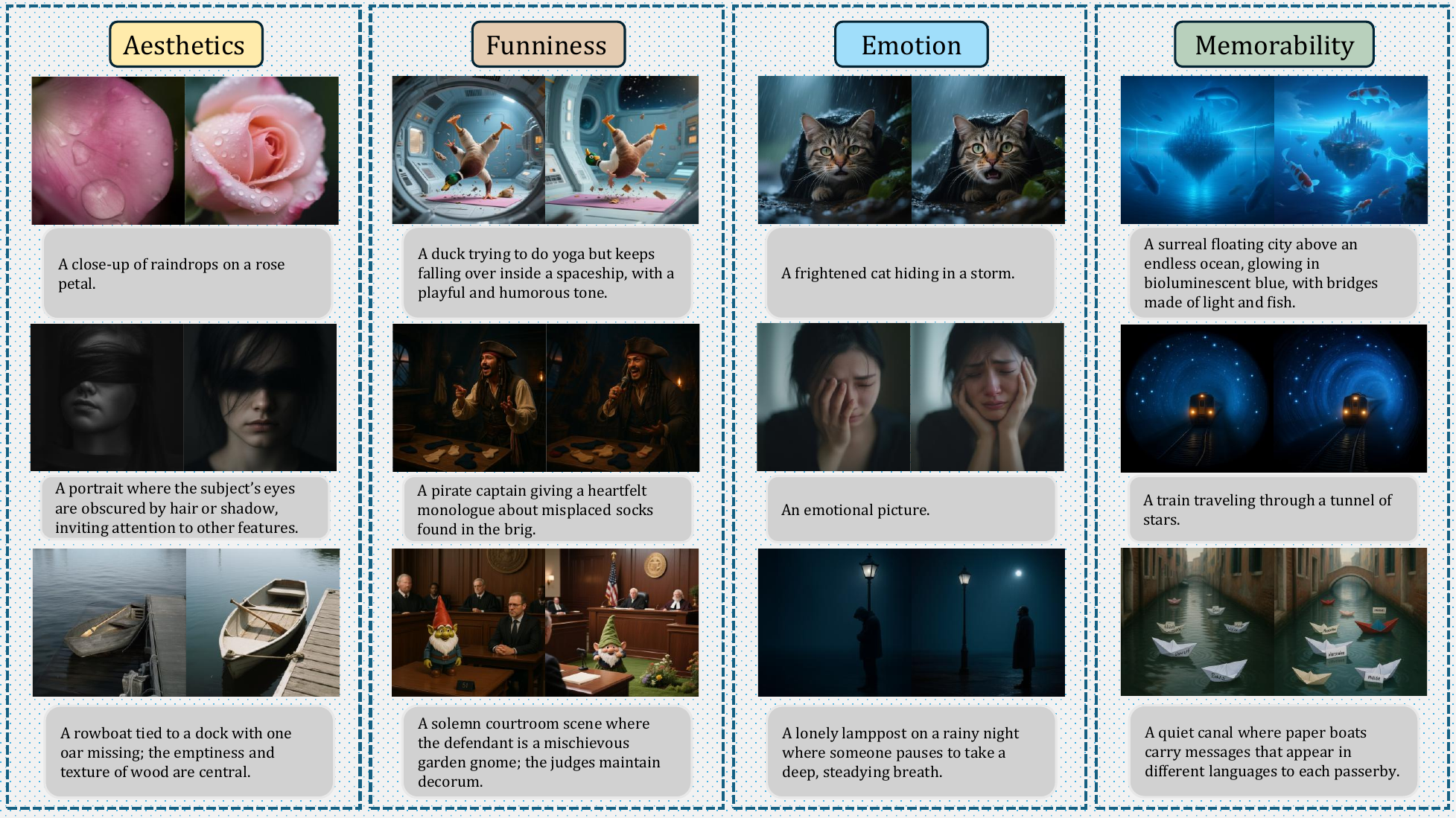}
    \caption{Qualitative comparison of images generated by the Qwen-Image pipeline using different LLM backbones (with the same prompt). The figure shows the effect of pretraining versus supervised fine-tuning (SFT) on image cognition properties. For each image pair, Left: Base model; right: SFT model. Generation prompts are shown under each image pair. We can see that images generated with our SFT MLLM backbone better demonstrate the cognitive cues embedded in the prompts.}
    \label{fig:qwenimage}
\end{figure*}

In this section, we demonstrate how cognition-aligned models differ from their base counterparts through an image generation pipeline using a consistent random seed for fair comparison. Specifically, we employ Qwen-Image \cite{wu2025qwenimagetechnicalreport}, a recently released image generation framework that adopts Qwen2.5-VL-7B-Instruct as its MLLM backbone for semantic encoding. Since this model also serves as the backbone for one of our SFT variants, we can seamlessly integrate it to compare image generations before and after fine-tuning. Uncurated visual examples are shown in Figure~\ref{fig:qwenimage}, highlighting the differences produced by the two LLM backbones.

\subsection{Prompt Preparation}
We split the prompts for image generation into five sets: \textbf{Aesthetics, Funniness, Emotional, Memorability} and \textbf{General}. For each set, we employed ChatGPT-5 to generate \textbf{100} text-to-image (T2I) prompts that demonstrate the properties of that type.

\subsection{Image Metrics Results}
For the \textbf{General-Prompting} images, we use some general human preferences and the text-image alignment metrics to analyze the impact of the cognition alignment by SFT. We used ClipScore \cite{hessel2022clipscorereferencefreeevaluationmetric}, Human Preference Score (HPS-v2) \cite{wu2023human}, LAION score \cite{laion_aesthetic_predictor}, and ImageReward score \cite{xu2023imagereward}. This is illustrated in Table \ref{tab:General QwenImage}. Among these, ImageReward is a human-trained reward model designed to assess prompt-to-image consistency and aesthetic preference. As shown in Table \ref{tab:General QwenImage}, all evaluation metrics improve after replacing the MLLM backbone with its SFT-aligned counterpart. Notably, ImageReward shows a substantial gain of +22.8\%, indicating that cognition alignment significantly enhances generation quality and strengthens alignment with human visual and aesthetic preferences. 

For the \textbf{Cognition-Prompting} image generation, we use cognition-specific models to evaluate the cognition scores of each image set (Table \ref{tab:Cognition QwenImage}). Models specifications can be found in \ref{subsec:experiment settings}, these specific models outperform most MLLMs across all cognitive dimensions (Table \ref{tab:CogIP-BenchResults}). 

We observe consistent improvements in cognition scores predicted by the post-trained models across all cognitive dimensions, confirming the effectiveness of post-training in enhancing MLLMs’ alignment with human cognition on visual inputs. Notably, the emotional cognition dimension exhibits the most significant gain (+19\%), suggesting that Qwen-Image has a stronger ability to perceive and interpret emotional cues present in the prompts.

\begin{table}[h]
  \centering
  \small
  \setlength{\tabcolsep}{3pt}
  \renewcommand{\arraystretch}{1.15}

  \resizebox{\columnwidth}{!}{
  \begin{tabular}{
    l
    S[table-format=1.3] 
    S[table-format=1.3] 
    S[table-format=2.3] 
    S[table-format=1.3] 
    S[table-format=1.3] 
  }
  \toprule
  \textbf{MLLM Backbone} & \textbf{ClipScore} & \textbf{HPS} & \textbf{PickScore} & \textbf{LAION} & \textbf{ImageReward} \\
  \midrule
  Qwen2.5-VL-7B (base)  & 0.330 & 0.267 & 21.973 & 6.765 & 0.878 \\
  Qwen2.5-VL-7B (ours)   & \cyan{0.332} & \cyan{0.277} & \cyan{22.134} & \cyan{6.796} & \cyan{1.078} \\
  \bottomrule
  \end{tabular}  
}   
  \caption{Image scores (General) for Qwen-Image--generated images with different MLLM backbones across various image-quality and human preference metrics.}
  \label{tab:General QwenImage}
   
\end{table}

  \begin{table}[h]
  \centering
  \small
  \setlength{\tabcolsep}{3pt}
  \renewcommand{\arraystretch}{1.15}

  \resizebox{\columnwidth}{!}{
  \begin{tabular}{
    l
    S[table-format=1.3] 
    S[table-format=1.3] 
    S[table-format=2.3] 
    S[table-format=1.3] 
  }
  \toprule
  \textbf{MLLM Backbone} & \textbf{Aesthetics} & \textbf{Funniness} & \textbf{Emotional} & \textbf{Memorability} \\
  \midrule
  Qwen2.5-VL-7B (base)  & 6.364 & 3.627 & 1.420 & 7.760  \\
  Qwen2.5-VL-7B (ours)   & \cyan{6.462} & \cyan{3.705} &\cyan{1.690} & \cyan{7.770}  \\
  \bottomrule
  \end{tabular} 
  }
  \caption{Image scores (Cognition) predicted by specific cognition-predicting models for Qwen-Image--generated images with different MLLM backbones.}
  \label{tab:Cognition QwenImage}

\end{table}

\subsection{User Study}
\begin{figure}[h]
    \centering
    \includegraphics[width=\linewidth]{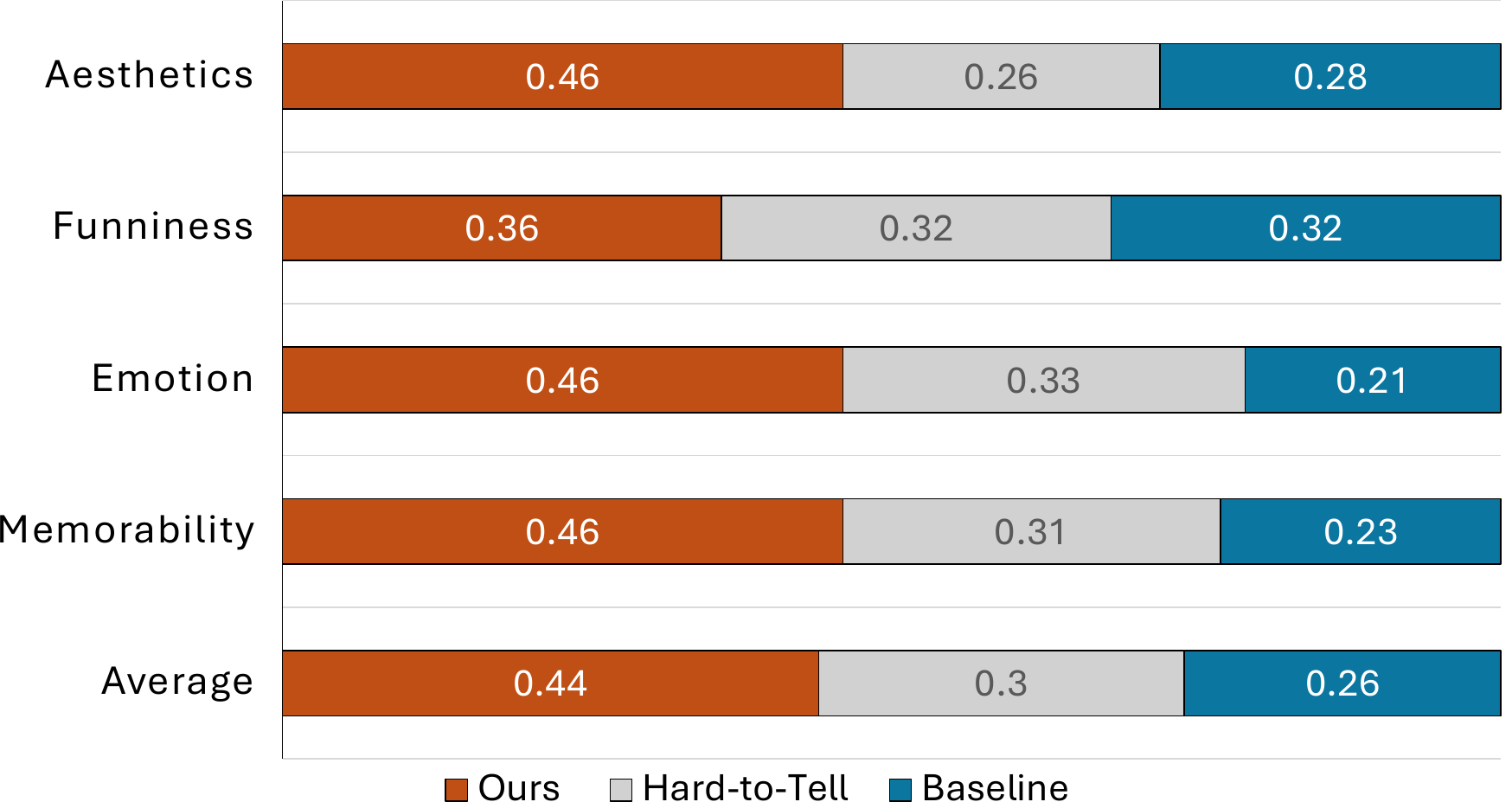}
    \caption{Preference percentages of images generated by Qwen-Image using the baseline MLLM backbone and our fine-tuned version.}
    \label{fig:user-study}
\end{figure}

Here, we present a user study evaluating how our cognition alignment pipeline improves human preference towards images generated by the Qwen-Image pipeline. We sampled 30 image pairs for each cognitive dimension, one generated by the baseline MLLM backbone and the other by our aligned version. Five volunteers were then asked to select their preferred image or choose ``Hard to Tell" if undecided. As shown in Figure~\ref{fig:user-study}, the images from our aligned model consistently achieved higher preference rates. On average, our model was preferred 1.7 times more frequently than the baseline.

\section{Discussions}
\subsection{Ablation Study} \label{subsec:ablation}
To examine the contribution of each component in our alignment strategy, we conduct an ablation study in four settings: (1) fine-tuning without the soft-label loss, (2) using a simplified prompting scheme that omits the “describe-then-predict” instruction format and directly asks the model to predict the score, (3) fine-tuning only the visual encoder while freezing the language model (LLM) and (4) fine-tuning only the LLM with the visual encoder frozen. We conduct this ablation study using Qwen2.5-VL-7B. We also include the average regression percentages of these fine-tuned models across four benchmarks: MMVP \cite{tong2024eyeswideshutexploring}, OCRBench \cite{Liu_2024}, MME \cite{fu2023mme} and MMMU \cite{yue2023mmmu} (each representing one task category \cite{tong_cambrian-1_2024}).
\\
Comparing the freeze-vision and freeze-LLM configurations (Table \ref{tab:ablation1}), we find that the vision encoder plays a pivotal role in image cognition tasks. The freeze-vision setting yields the weakest alignment across all cognitive dimensions, whereas the freeze-LLM setting achieves slightly better performance, particularly on the aesthetics and funniness dimensions. This observation contrasts with common fine-tuning practices in downstream multimodal tasks, where freezing the visual encoder is often preferred for stability and efficiency. Our findings suggest that the visual subsystem of MLLMs contributes more substantially to human-aligned image cognition. However, the freeze-LLM SFT setting leads to a significant decline in general multimodal reasoning tasks, indicating that a balanced optimization between the visual and language components remains crucial for maintaining overall model capability.

\begin{table}[h]
\centering
\small
\setlength{\tabcolsep}{3pt}
\renewcommand{\arraystretch}{1.1}

\resizebox{\columnwidth}{!}{
\begin{tabular}{
l
S[table-format=1.3]
S[table-format=1.3]
S[table-format=1.3]
S[table-format=1.3]
S[table-format=1.3]
}
\toprule
\textbf{Model} &
\textbf{Aesthetics} &
\textbf{Funniness} &
\textbf{Emotional} &
\textbf{Memorability}&
\textbf{Regression}\\
\midrule

SFT
& 3.841 & 7.022 & \pos{5.775} &\pos{15.500} & -0.01\% \\

Freeze-Vision
& \neg{5.923} & \neg{7.488} & \neg{7.634} & \neg{32.290} &-0.74\% \\

Freeze-LLM
& \pos{2.511} & \pos{5.031} & 6.427 & \nat{23.440} & -5.37\%\\

\bottomrule
\end{tabular}
}

\caption{Ablation results (MSE only) across four cognitive dimensions, comparing three frozen-encoder configurations of Qwen2.5-VL-7B. The SFT setting is our baseline alignment method. Regression represents the average performance degradation percentage on other multimodal reasoning benchmarks.}
\label{tab:ablation1}
\end{table}

For the ablation of prompting techniques and soft-label loss (Table \ref{tab:ablation2}), we observe that incorporating the soft-label loss generally enhances alignment across all cognitive dimensions, demonstrating its effectiveness in numeric prediction tasks. Finally, the proposed prompting strategy yields the most substantial gains, improving cognitive alignment notably across all dimensions.

\begin{table}[h]
\centering
\small
\setlength{\tabcolsep}{3pt}
\renewcommand{\arraystretch}{1.1}

\resizebox{\columnwidth}{!}{
\begin{tabular}{
l
S[table-format=1.3]
S[table-format=1.3]
S[table-format=1.3]
S[table-format=1.3]
S[table-format=1.3]
}
\toprule
\textbf{Model} &
\textbf{Aesthetics} &
\textbf{Funniness} &
\textbf{Emotional} &
\textbf{Memorability }&
\textbf{Regression}\\
\midrule
SFT
& \pos{1.526} & \pos{2.037} & \pos{1.750} & \pos{3.010} & -0.01\% \\

Simple-Prompt
& \neg{5.570} & \neg{8.612} & \neg{6.678} & \neg{13.200} & -0.38\%\\

No Soft-Label
& 3.922 & 6.383 & 5.956 & \nat{14.320} & +0.03\%\\
\bottomrule
\end{tabular}
}

\caption{Ablation results (MSE only) demonstrating the effect of our prompting techniques and soft-label loss on our baseline alignment method using Qwen2.5-VL-7B.}
\label{tab:ablation2}
\end{table}

\subsection{Cognition-Alignment with RL} \label{sec:rl}
Can RL methods further align MLLMs with human cognition? To investigate this, we apply Group Relative Policy Optimization (GRPO) \cite{shao2024deepseekmathpushinglimitsmathematical} on our benchmark, fine-tuning the MLLM with a reward signal based on the numerical proximity between the model’s predicted score and the ground-truth (GT). Unlike supervised fine-tuning, which relies solely on static label supervision, GRPO introduces an adaptive learning signal that encourages models to iteratively refine their predictions toward human-like evaluations. Although, in our benchmark, the GT responses do not include explicit reasoning traces, the reward design implicitly instills numerical alignment awareness into the learning process, encouraging more human-consistent judgment calibration. Empirically, this reinforcement-based method yields improved alignment on aesthetics and emotional cognition dimensions compared to standard SFT (as shown in Table \ref{tab:grpo}), though it exhibits some degradation in broader multimodal reasoning tasks (with the same benchmark setup as in Section~\ref{subsec:ablation}). These results highlight the promise of policy optimization techniques for capturing human cognitive patterns in multimodal alignment.
 
\begin{table}[h]
\centering
\small
\setlength{\tabcolsep}{3pt}
\renewcommand{\arraystretch}{1.1}

\resizebox{\columnwidth}{!}{
\begin{tabular}{
l
S[table-format=1.3]
S[table-format=1.3]
S[table-format=1.3]
S[table-format=1.3]
S[table-format=1.3]
}
\toprule
\textbf{Model} &
\textbf{Aesthetics} &
\textbf{Funniness} &
\textbf{Emotional} &
\textbf{Memorability}&
\textbf{Regression}\\
\midrule

SFT
& 3.841 & 7.022 & 5.775 & \nat{15.500} & -0.01\% \\

GRPO
& \pos{1.760} & \neg{7.922} & \pos{4.514} & \neg{16.600} & -3.38\% \\

\bottomrule
\end{tabular}
}

\caption{MSE results of the \textbf{GRPO} fine-tuning using the Qwen2.5-VL-7B model for each dimension (lower is better).}
\label{tab:grpo}
\end{table}

\section{Conclusion}
Our work demonstrates that the subjective, cognitive dimension of visual understanding is no longer beyond the reach of MLLMs. By systematically measuring and then bridging the gap between machine and human perception, we have shown that these models can be taught a semblance of human-like taste and intuition. Furthermore, this learned alignment is not only an end in itself but a transferable capability that unlocks more intuitive and human-centric creative applications--a step towards more deeply human-aligned AI. We hope this work serves as a step toward creating AI systems that can not only see but also feel the world as human do.

\clearpage

{
    \small
    \bibliographystyle{ieeenat_fullname}
    \bibliography{main}
}


\clearpage            
\onecolumn            
\appendix             
\setcounter{figure}{0}
\setcounter{table}{0}
\setcounter{equation}{0}
\renewcommand{\thefigure}{S\arabic{figure}}
\renewcommand{\thetable}{S\arabic{table}}
\renewcommand{\theequation}{S\arabic{equation}}

\section*{\huge Appendices}
\vspace{2em}

\section{Detailes of Results on Other Benchmarks}
This section presents the detailed performance of the three MLLMs on 12 benchmarks spanning four task categories, evaluated both before and after SFT on our CogIP-Bench:

\begin{table*}[h]
\centering
\small
\setlength{\tabcolsep}{3.5pt}
\renewcommand{\arraystretch}{1.1}

\begin{adjustbox}{width=\textwidth,center}
\begin{tabular}{lcccccccccccccccc}
\toprule
\textbf{Model} &
\multicolumn{4}{c}{\textbf{Vision-Centric}} &
\multicolumn{4}{c}{\textbf{OCR}} &
\multicolumn{4}{c}{\textbf{General}} &
\multicolumn{4}{c}{\textbf{Knowledge}} \\
\cmidrule(lr){2-5} \cmidrule(lr){6-9} \cmidrule(lr){10-13} \cmidrule(lr){14-17}

& \textbf{MMVP} & \textbf{RealWorldQA} & \textbf{CVBench} & \textbf{Avg} &
\textbf{OCRBench} & \textbf{DocVQA} & \textbf{ChartQA} & \textbf{Avg} &
\textbf{MME} & \textbf{MMBench} & \textbf{SeedBench} & \textbf{Avg} &
\textbf{MMMU} & \textbf{MathVista} & \textbf{AI2D} & \textbf{Avg} \\
\midrule

Qwen2.5-Base &
0.780 & 0.686 & 0.755 & \textbf{0.740} &
0.884 & 0.949 & 86.2 & \textbf{0.898} &
0.849 & 0.833 & 0.772 & \textbf{0.818} &
0.577 & 0.681 & 0.847 & \textbf{0.702} \\

Qwen2.5B-SFT &
0.777 & 0.693 & 0.743 & \neg{\textbf{0.738}}
&0.882 & 0.951 &0.856 &\neg{\textbf{0.896}}
&0.847 & 0.842 & 0.771 &\pos{\textbf{0.820}}
&0.581 & 0.681 & 0.847 & \pos{\textbf{0.703}} \\
\midrule
Gemma3-Base &
0.723 & 0.607 & 0.692 & \textbf{0.674} &
0.702 & 0.722 & 0.372 & \textbf{0.599} &
0.765 & 0.749 & 0.692 & \textbf{0.735} &
0.557 & 0.551 & 0.790 & \textbf{0.633} \\

Gemma3-SFT &
0.720 & 0.617 & 0.716 & \pos{\textbf{0.684}} &
0.754 & 0.808 & 0.391 & \pos{\textbf{0.651}} &
0.776 & 0.739 & 0.719 & \pos{\textbf{0.745}} &
0.564 & 0.580 & 0.823 & \pos{\textbf{0.656}} \\
\midrule
Llama-3.2-Base &
0.717 & 0.630 & 0.710 & \textbf{0.686} &
0.755 & 0.825 & - & \textbf{0.790} &
0.705 & 0.673 & 0.731 & \textbf{703} &
0.473 & 0.482 & 0.776 & \textbf{0.577} \\

Llama-3.2-SFT &
0.707 & 0.648 & 0.701 & \neg{\textbf{0.685}} &
0.741 & 0.845 & - & \pos{\textbf{0.793}} &
0.696 & 0.672 & 0.732 & \neg{\textbf{0.700}} &
0.494 & 0.482 & 0.778 & \pos{\textbf{0.585}} \\
\bottomrule
\end{tabular}
\end{adjustbox}

\caption{Benchmark comparison across multiple models with section averages (Vision-Centric, OCR, General, Knowledge). \postext{Green} entry color indicates an improvement on the average benchmark result on a certain task category of a particular MLLM, while the \negtext{red} entry color indicates a regression.}
\label{tab:other benchmark}
\end{table*}

\section{Training Details}
In this section, we report all hyperparameters used in our experiments—including the soft-label parameters—for training the three model variants, as summarized in Table~\ref{tab:hyperparams_all}.

\begin{table}[h]
    \centering
    \small
    \begin{tabular}{lccc}
        \toprule
        \textbf{Hyperparameter} & \textbf{Qwen2.5} & \textbf{Gemma3} & \textbf{Llama3.2} \\
        \midrule
        LoRA Rank & 64 & 64 & 64 \\
        LoRA $\alpha$ & 64 & 64 & 64 \\
        LoRA Dropout & 0.05 & 0.05 & 0.05 \\
        LoRA Target & all & all & all \\
        GPU & 4 $\times$ A40 & 4 $\times$ A40 & 4 $\times$ A40 \\
        Batch Size & 4 & 4 & 4 \\
        Gradient Accumulation Steps & 8 & 8 & 8 \\
        Warmup Ratio & 0.03 & 0.03 & 0.03 \\
        Learning Rate & 3e-5 & 1e-5 & 9e-5 \\
        Learning Rate Scheduler & Cosine & Cosine & Cosine \\
        Unfreeze Vision Tower & True & True & True \\
        \midrule
        \multicolumn{4}{c}{\textbf{Soft Label}} \\
        \midrule
        $\eta$ & 0.15 & 0.15 & 0.08 \\
        $\lambda$ & 2.0 & 2.0 & 2.0 \\
        \bottomrule
    \end{tabular} 
    \caption{Hyperparameters specification for training Qwen2.5-VL-7B, Gemma3-12B-it, and Llama-3.2-11B-VI models.}
    \label{tab:hyperparams_all}
\end{table}

\section{Example fo User Study}
This section presents example image pairs used in the user study assessing preferences for images generated by Qwen-Image with either the default or SFT-aligned MLLM backbones. For each cognitive dimension, 30 image pairs were randomly sampled and arranged side-by-side, after which five volunteers were asked to select the preferred image—without knowing which model produced it—or choose “Hard to Tell” when undecided.
\begin{figure}[H]
    \centering
    \vspace{-0.35em}
    \begin{subfigure}[t]{0.42\linewidth}
        \centering
        \includegraphics[width=\linewidth]{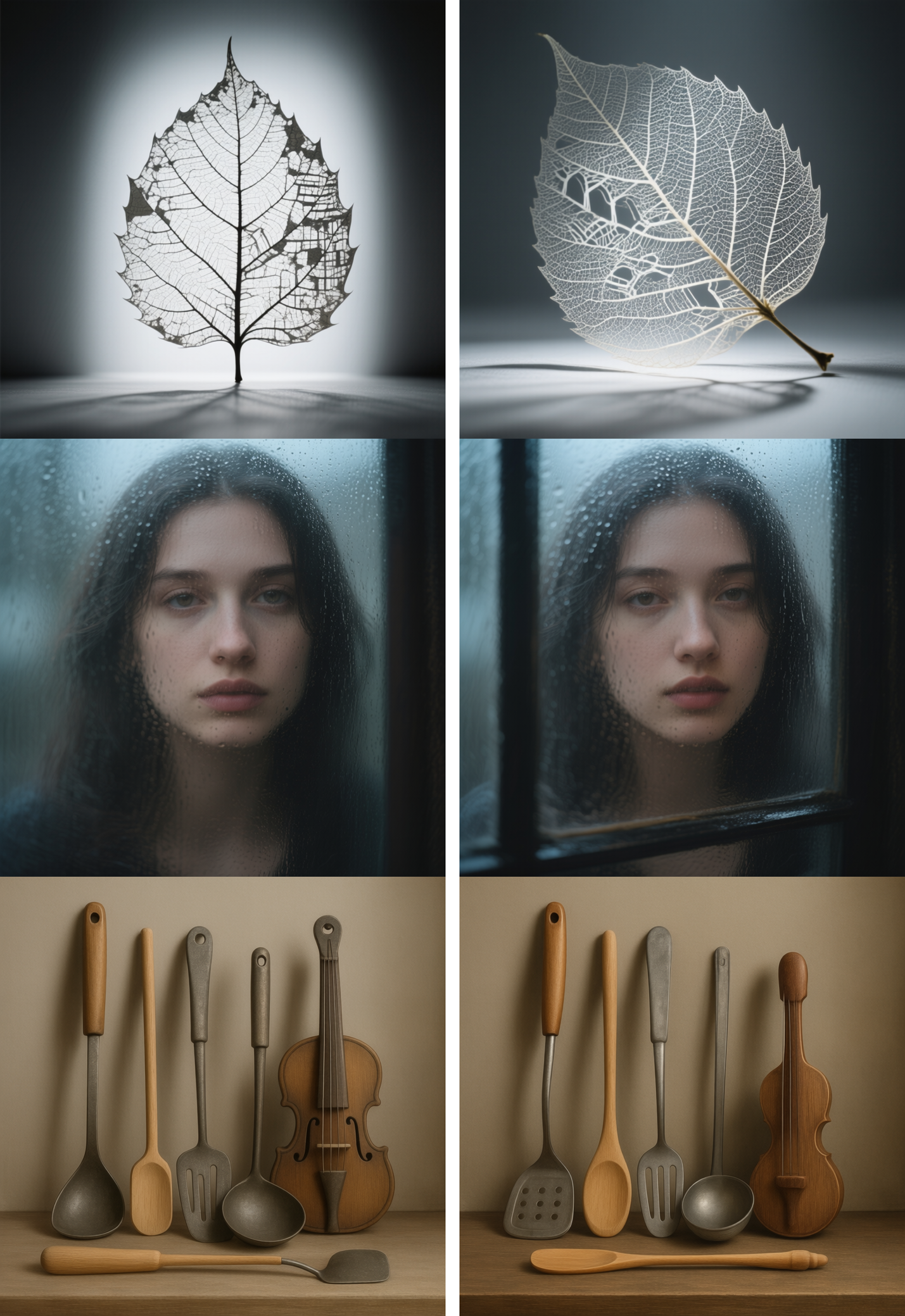}
        \caption{Aesthetics}
        \label{fig:Aesthetics}
    \end{subfigure}
    \hspace{0.5em}
    \begin{subfigure}[t]{0.42\linewidth}
        \centering
        \includegraphics[width=\linewidth]{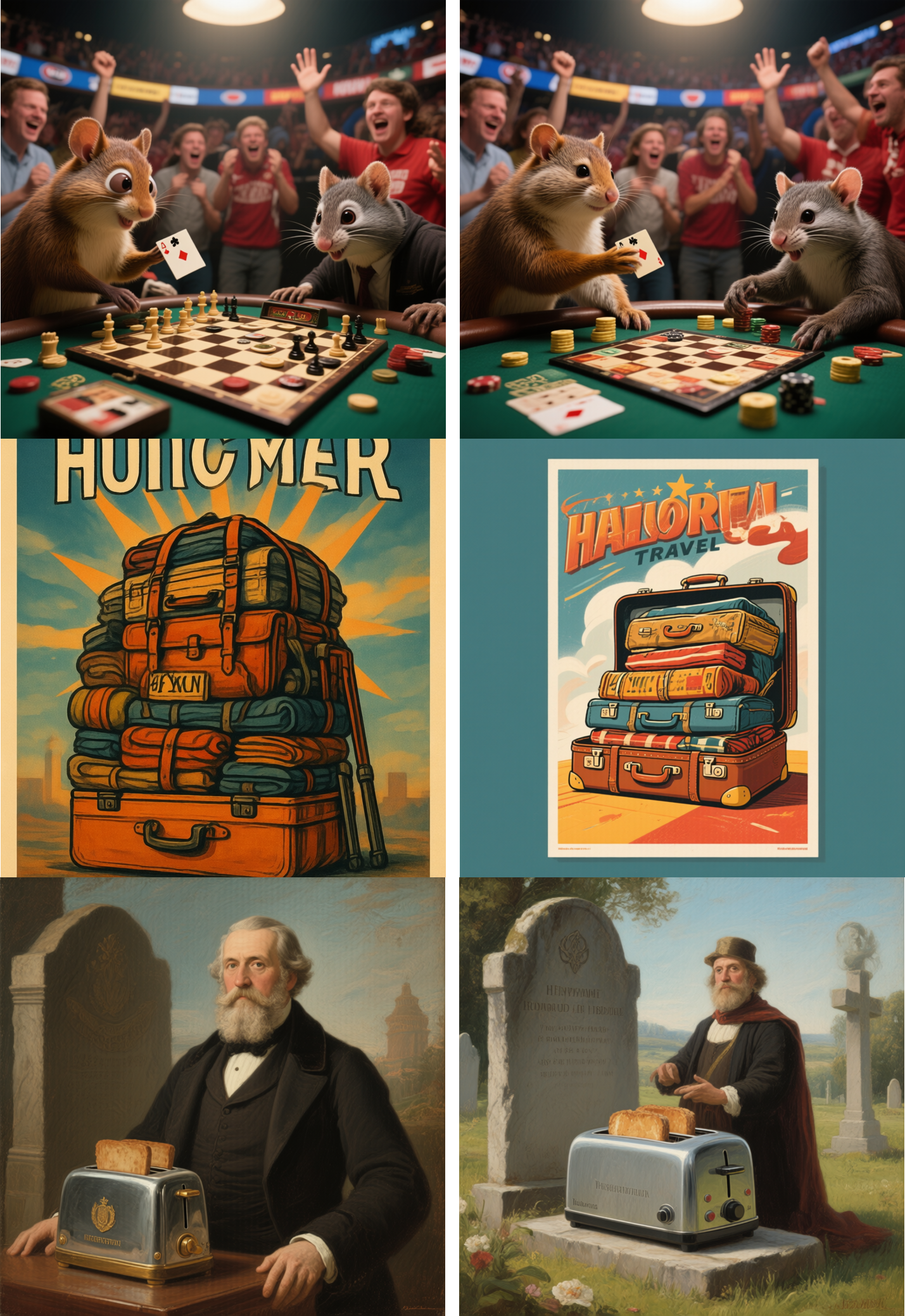}
        \caption{Funniness}
        \label{fig:Funniness}
    \end{subfigure}
    
    \vspace{1em}

    \begin{subfigure}[t]{0.42\linewidth}
        \centering
        \includegraphics[width=\linewidth]{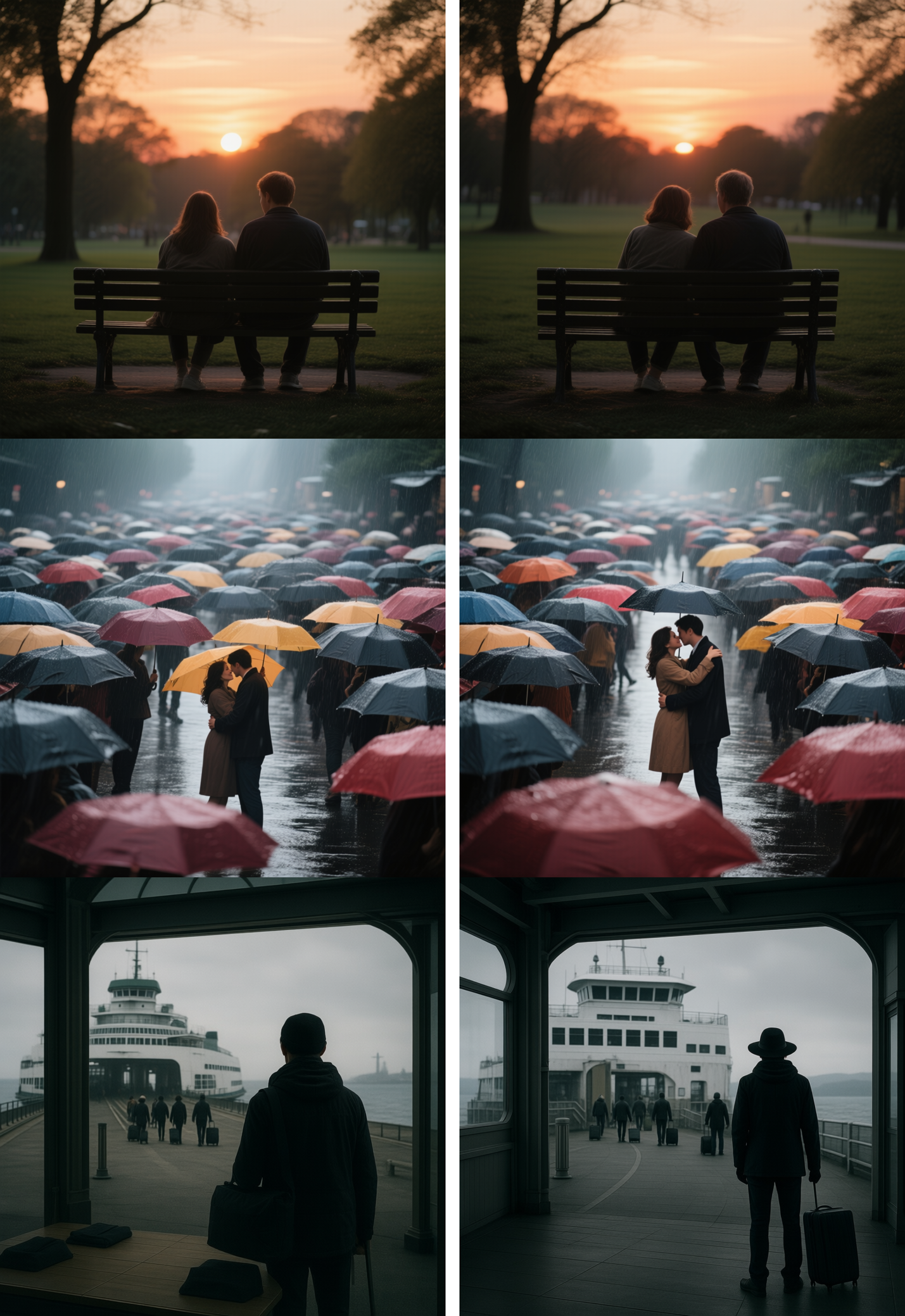}
        \caption{Emotional Valence}
        \label{fig:Emotional Valence}
    \end{subfigure}
    \hspace{0.5em}
    \begin{subfigure}[t]{0.42\linewidth}
        \centering
        \includegraphics[width=\linewidth]{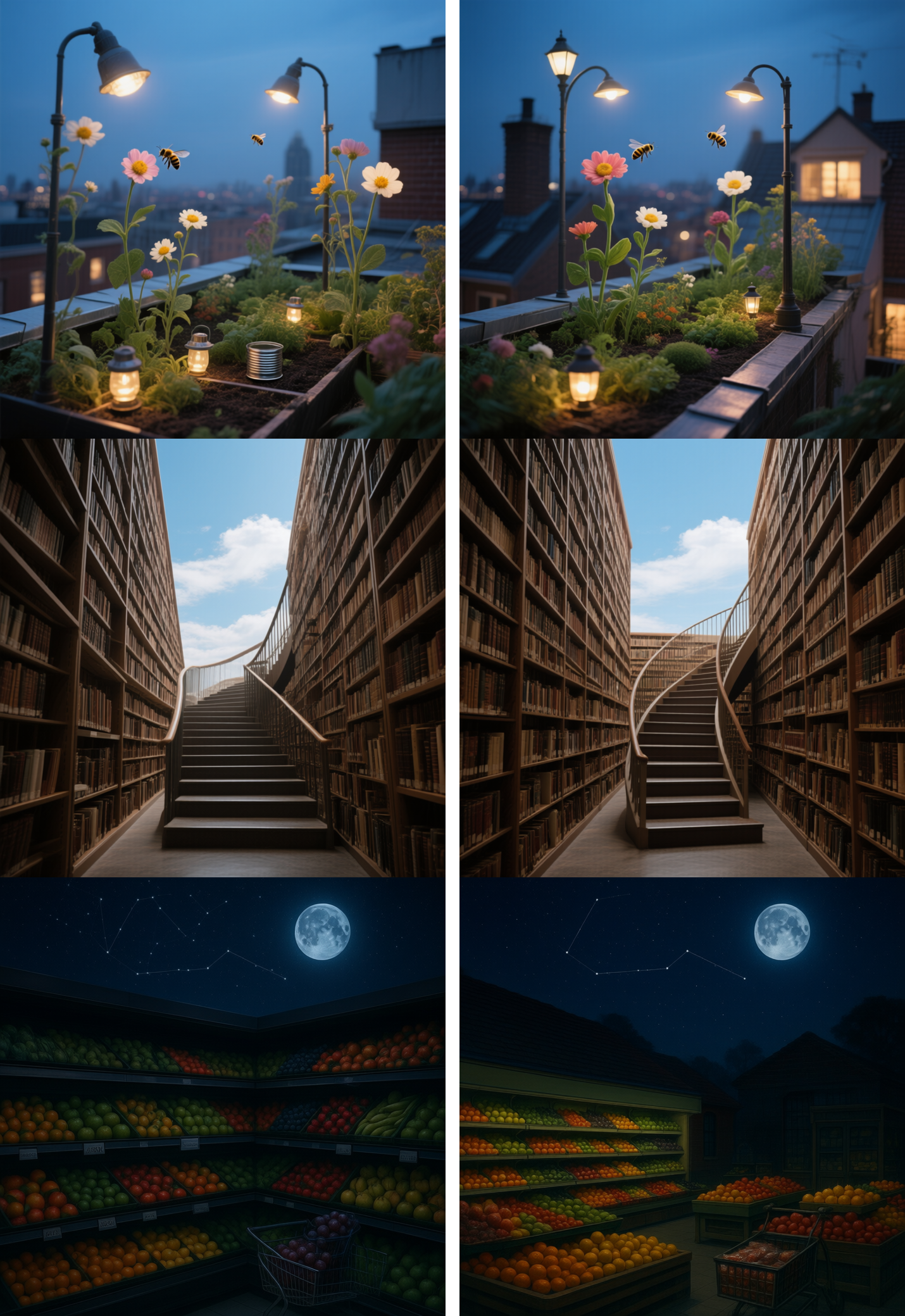}
        \caption{Memorability}
        \label{fig:Memorability}
    \end{subfigure}
    
    \caption{Demonstration of the user study set up, each pair is generated with the same prompt using Qwen-Image using consistent seed.}
    \label{fig:usr_study}
\end{figure}

\section{Examples of SFT Dataset}
\begin{figure}[H]
    \centering
    \includegraphics[width=0.93\linewidth]{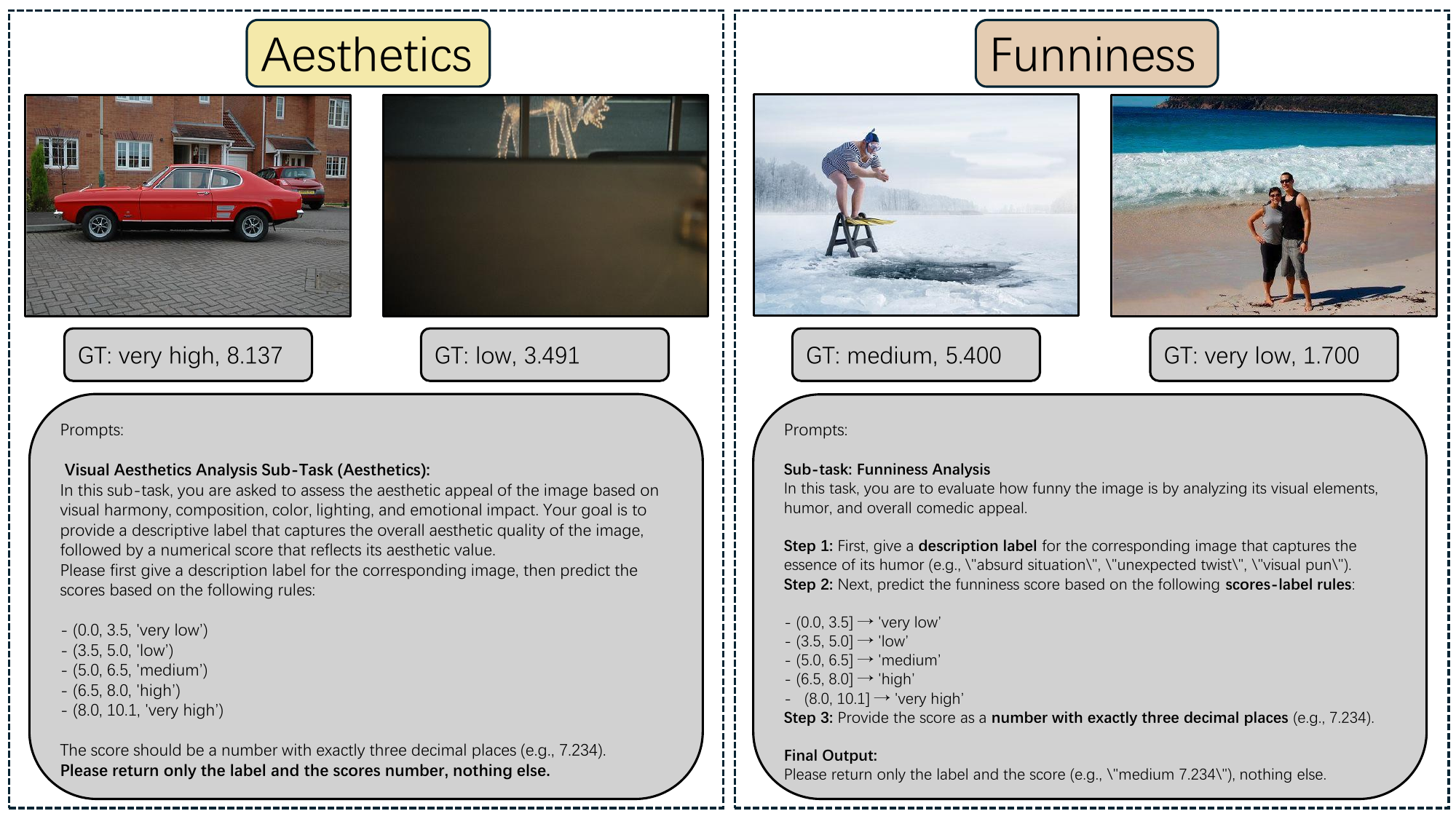}
    \caption{Examples of CogIP-Bench samples for Aesthetics and Funniness}
    \label{fig:bench1}
\end{figure}

\begin{figure}[H]
    \centering
    \includegraphics[width=0.93\linewidth]{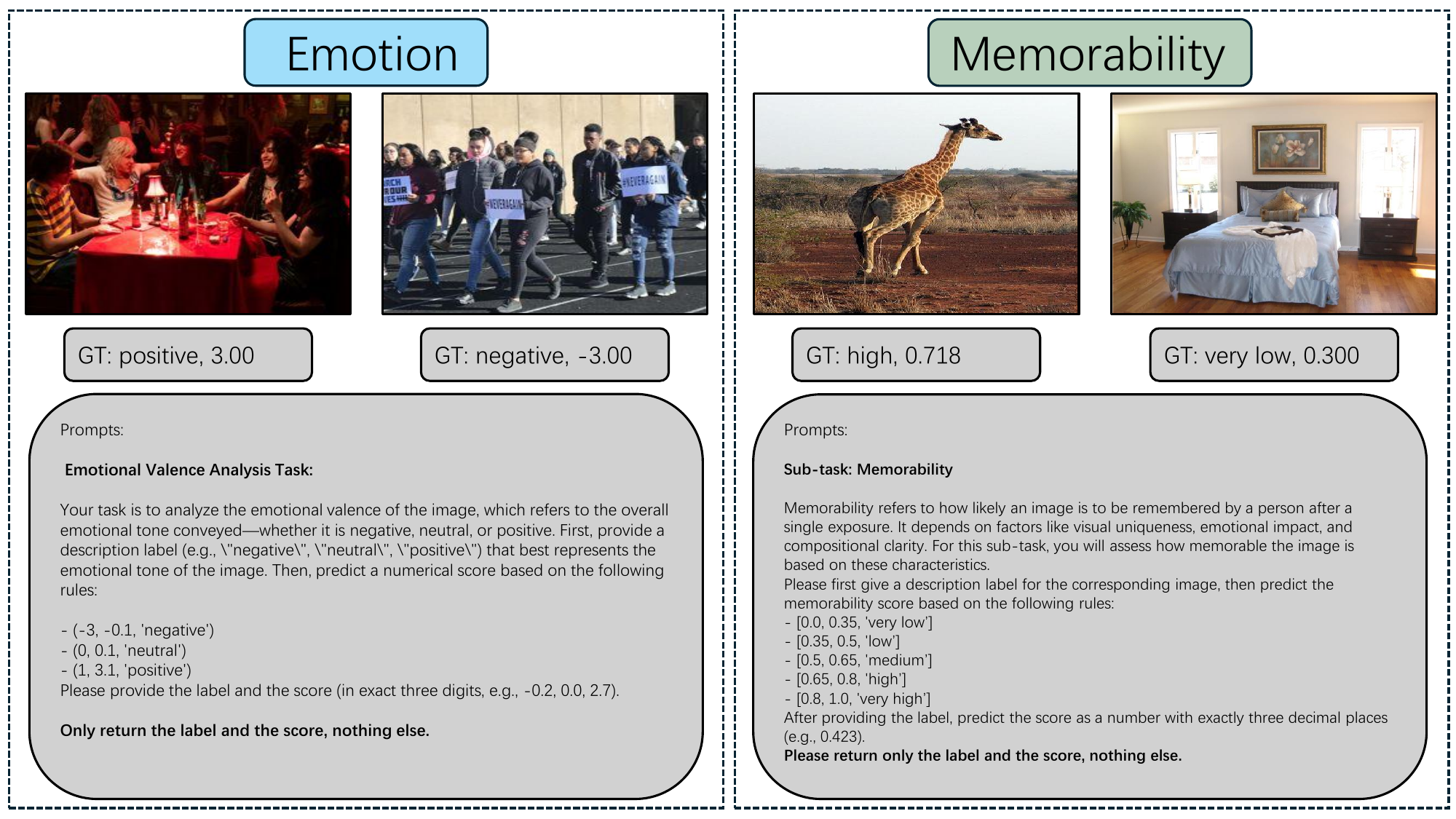}   
    \caption{Examples of CogIP-bench samples of Emotional Valence and Memorability.}
    \label{fig:bench2} 
\end{figure}

In this section, we provide example samples from our constructed CogIP-Bench dataset used for supervised fine-tuning, as shown in Figures~\ref{fig:bench1} and \ref{fig:bench2}. The dataset spans four cognitive dimensions.

\clearpage

\section{Prompts}

\begin{figure}[H]
    \centering
    \includegraphics[width=0.99\linewidth]{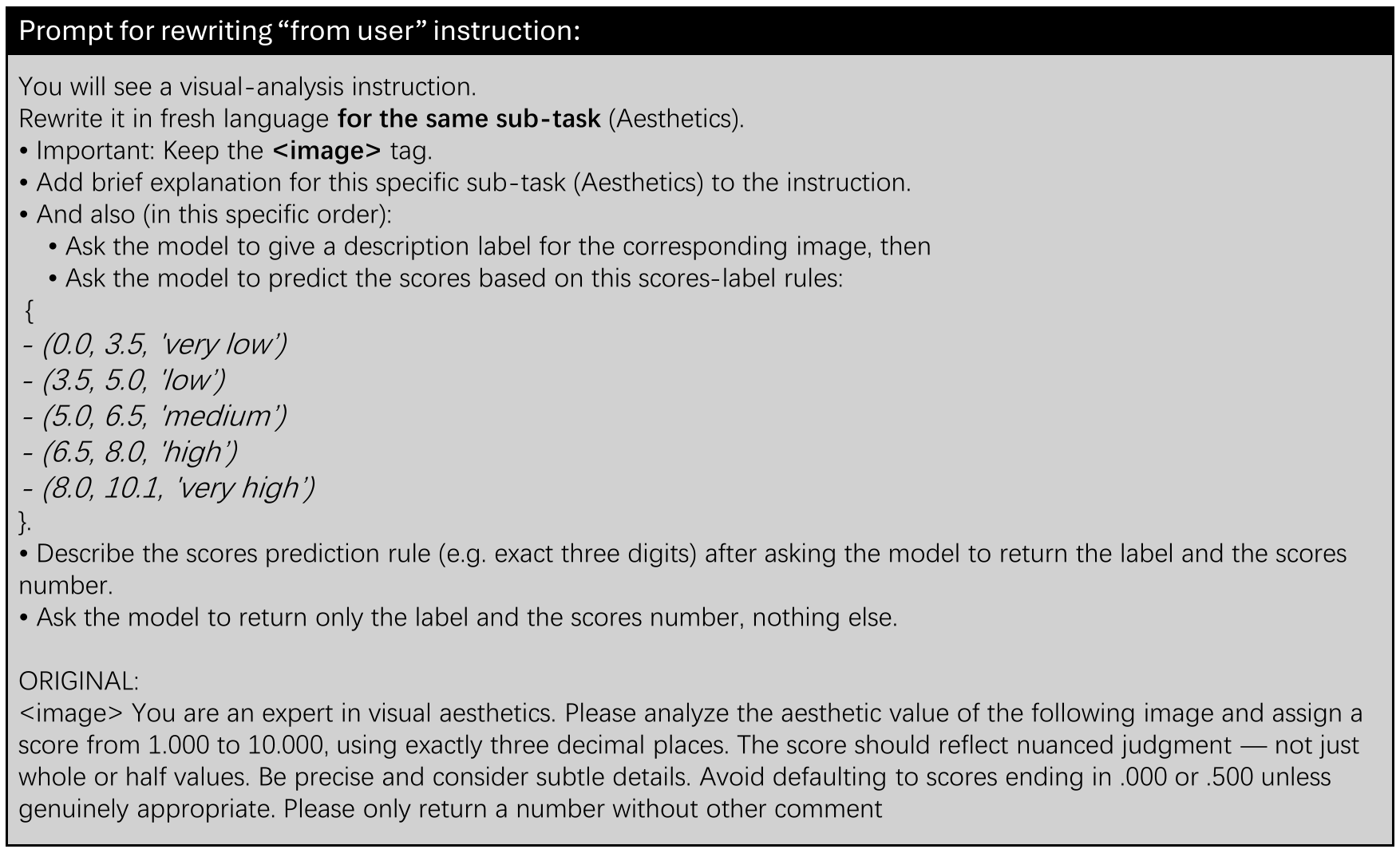}
    \caption{Prompts for rewriting ``from human" instruction in the dataset.}
    \label{fig:prompt1}
\end{figure}

\begin{figure}[h]
    \centering
    \includegraphics[width=0.99\linewidth]{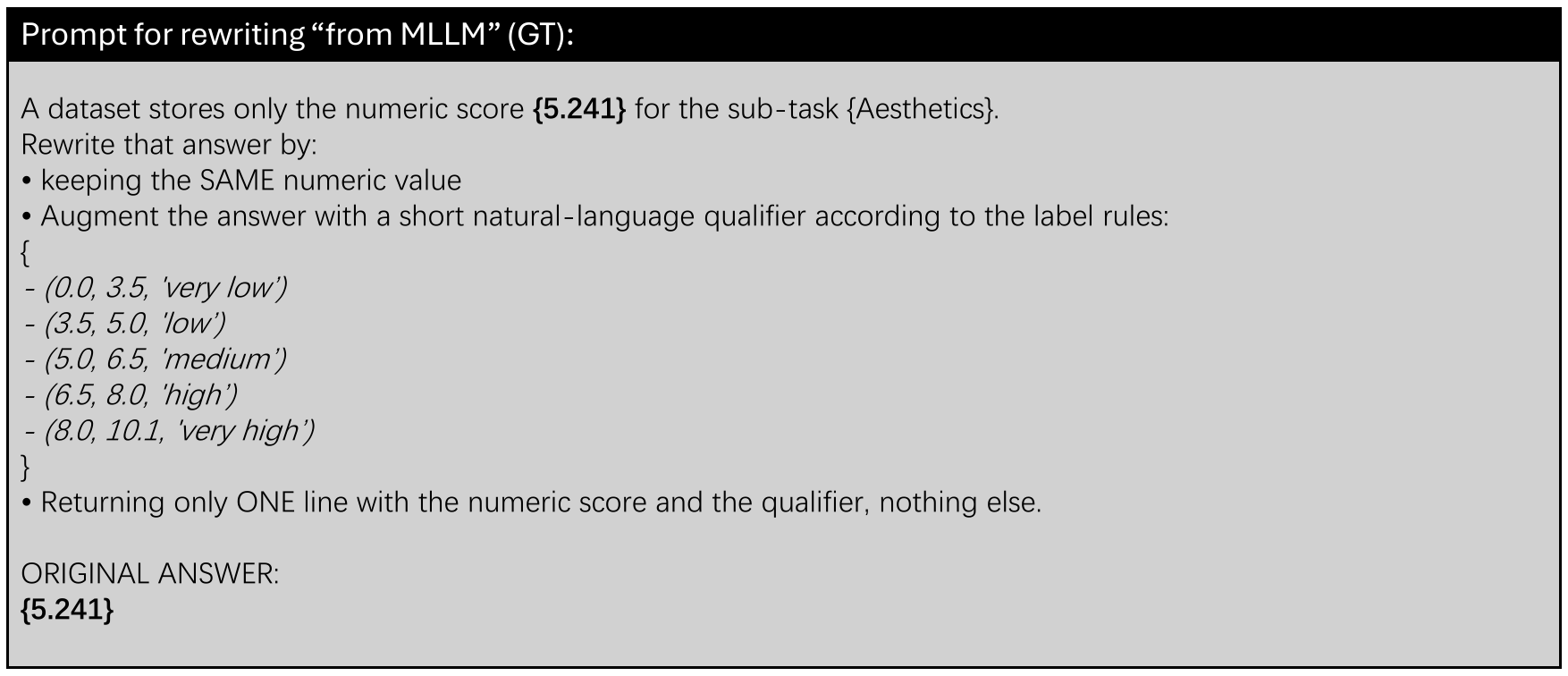}
    \caption{Prompts for rewriting ``from gpt" GT in the dataset.}
    \label{fig:prompt2}
\end{figure}

\begin{figure}[h]
    \centering
    \includegraphics[width=0.99\linewidth]{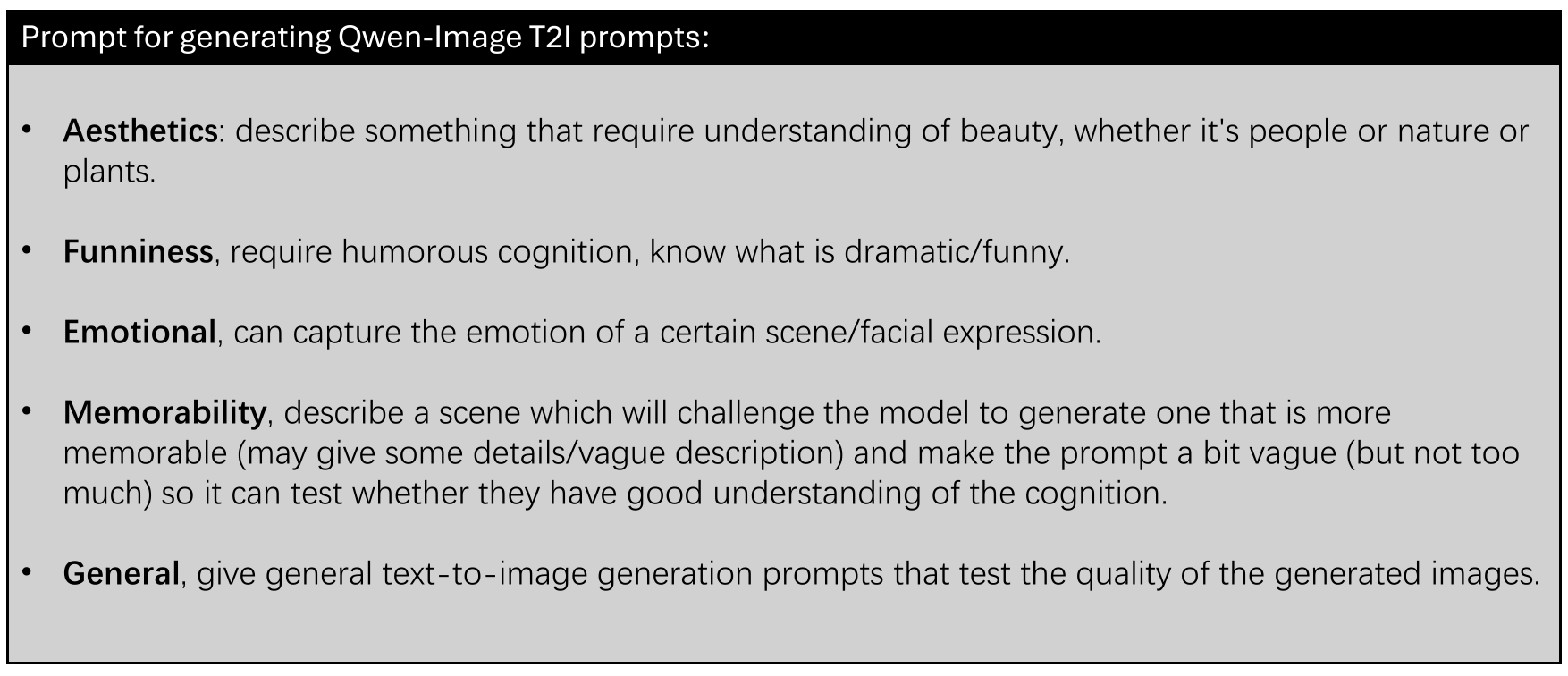}
    \caption{Prompt to generate T2I generation prompts for Qwen-Image}
    \label{fig:prompt3}
\end{figure}

In this section, we systematically present all prompts used to construct the CogIP-Bench dataset and to evaluate the SFT effects on Qwen-Image. Specifically:

\begin{itemize}
\item Figure~\ref{fig:prompt1} illustrates the prompt used to rewrite the original simple instruction prompts for the dataset—shown here using aesthetics as an example—including the reformulation of the ``from human'' instruction. \item Figure~\ref{fig:prompt2} presents the prompt employed to rewrite the ``from gpt'' ground-truth responses.
\item Figure~\ref{fig:prompt3} provides the prompts used to generate text-to-image instructions designed to elicit specific cognitive attributes, along with a general-purpose version for comparison.
\end{itemize}

\end{document}